\pdfoutput=1

\documentclass[11pt]{article}

\usepackage[final]{acl}

\usepackage{times}
\usepackage{latexsym}

\usepackage[T1]{fontenc}

\usepackage[utf8]{inputenc}

\usepackage{microtype}

\usepackage{inconsolata}

\usepackage{graphicx}

\usepackage{fontawesome}
\usepackage{hyperref}
\usepackage{siunitx}
\usepackage{kotex}
\usepackage{multirow}
\usepackage{amsmath}
\usepackage{amsfonts}
\usepackage{graphicx}
\usepackage{cleveref}
\usepackage{graphicx}
\usepackage{dblfloatfix} 
\usepackage{subcaption}
\usepackage{booktabs}
\usepackage{float}
\usepackage[toc,page]{appendix}
\usepackage[export]{adjustbox}
\usepackage{colortbl}
\crefformat{section}{\S#2#1#3}
\crefformat{subsection}{\S#2#1#3}
\crefformat{subsubsection}{\S#2#1#3}
\usepackage[normalem]{ulem}

\title{Revisiting the Impact of Pursuing Modularity for Code Generation}

\author{Deokyeong Kang$^{\dagger}$, Ki Jung Seo$^{\dagger}$, Taeuk Kim$^*$ \\
Department of Computer Science, Hanyang University, Seoul, Republic of Korea \\
{\tt \{rkdejrdud88,tjrlwjd1,kimtaeuk\}@hanyang.ac.kr}}

\newcommand{\astfootnote}[2]{
    \let\oldthefootnote=\thefootnote
    \setcounter{footnote}{#1}
    \renewcommand{\thefootnote}{\fnsymbol{footnote}}\footnotetext{#2}
    \let\thefootnote=\oldthefootnote
}

\begin{document}
\maketitle
\begin{abstract}
Modular programming, which aims to construct the final program by integrating smaller, independent building blocks, has been regarded as a desirable practice in software development.
However, with the rise of recent code generation agents built upon large language models (LLMs), a question emerges: is this traditional practice equally effective for these new tools?
In this work, we assess the impact of modularity in code generation by introducing a novel metric for its quantitative measurement.
Surprisingly, unlike conventional wisdom on the topic, we find that modularity is not a core factor for improving the performance of code generation models.
We also explore potential explanations for why LLMs do not exhibit a preference for modular code compared to non-modular code. Our code is available at {\faGithub} \href{https://github.com/HYU-NLP/Revisiting-Modularity}{https://github.com/HYU-NLP/Revisiting-Modularity}.
\end{abstract}

\astfootnote{2}{Equal contribution. $^*$Corresponding author.}
\setcounter{footnote}{0}

\section{Introduction}

With recent advances in the capabilities of large language models (LLMs; \citealp{openai2024gpt4,geminiteam2024gemini}; \textit{inter alia}), their application areas have expanded beyond simple text-based tasks. 
Among these, coding assistants are becoming practically essential for programmers, enhancing their efficiency through tasks such as natural language to code (NL2Code) generation.


Similar to other use cases of LLMs, coding assistants are typically utilized in zero- or few-shot manners.
The problem is that as the length of code is usually much longer than that of a sentence, the number of code examples available for each run is strictly limited.
Furthermore, the same functionality can be represented with different forms of code, making it challenging for users to select a proper example for a target task.
The diversity of code formats also poses challenges in fine-tuning setups, as constructing an appropriate training dataset becomes non-trivial.
It is thus important to understand what characteristics of the code provided to the agents contribute to their final performance of such models.
Among the many possible properties that influence the characteristics of code snippets, this work investigates the impact of \textbf{code modularity} on the performance of LLMs for NL2Code generation.

\begin{figure}[t!]
\centering
    \includegraphics[width=\columnwidth]{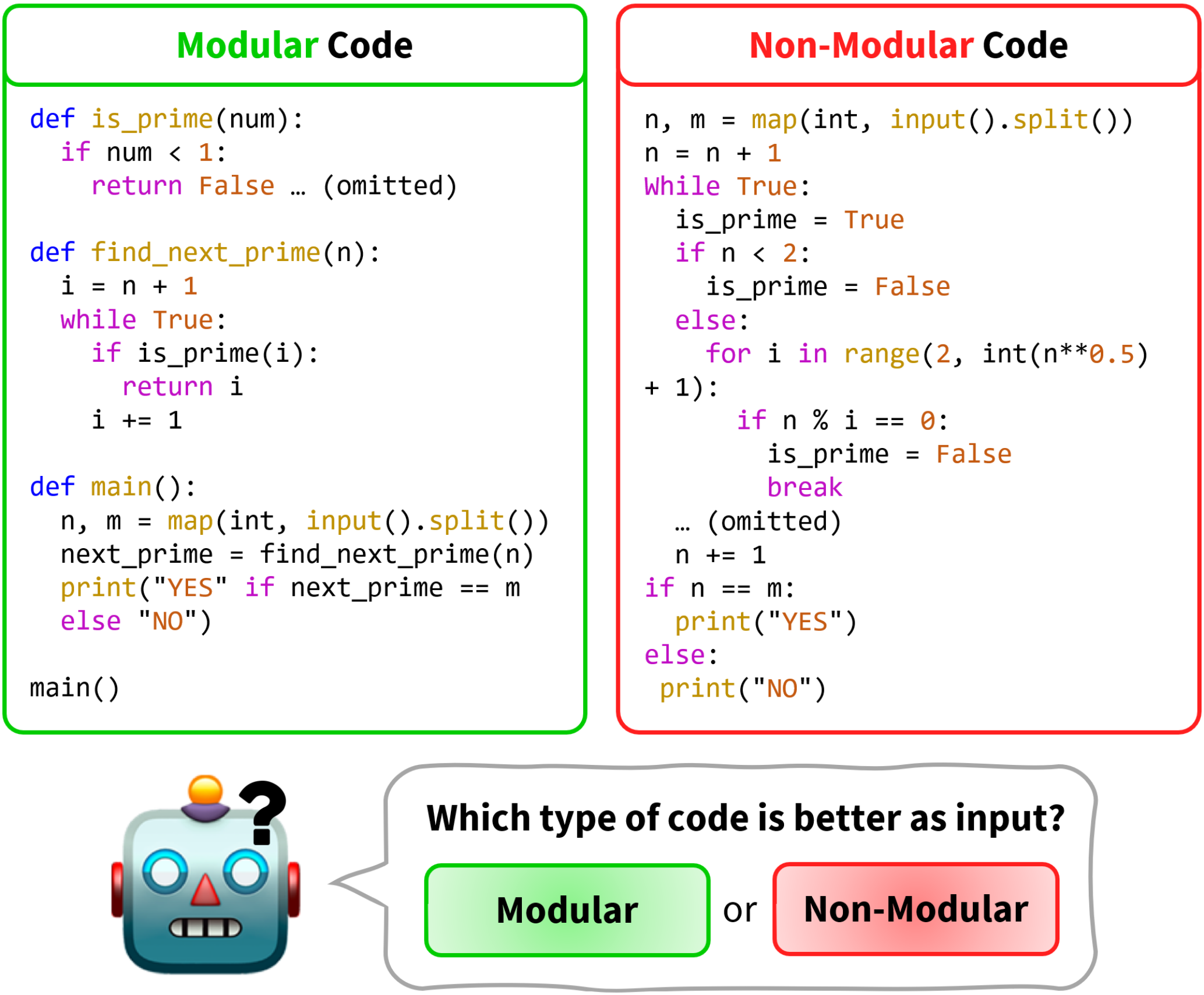}
    \caption{In this work, we address the following research question: Given modular and non-modular code snippets with identical functionality, which code type more effectively enhances performance in code generation when used as input for code language models?}
    \label{fig:main}
\end{figure}

Modular programming, the practice of building software with independent components, has long been considered a cornerstone of good software development. 
While this paradigm facilitates desirable properties of code for \textit{human programmers}, such as reusability, readability, and maintainability, it remains an open question whether it offers the same level of effectiveness for \textit{LLMs}.
 
Notably, \citet{jain2024llmassisted} argued that leveraging a set of modular functions can improve code generation accuracy for both in-context learning (ICL) and fine-tuning.
As it is not trivial to guarantee the modularity of each code snippet, the authors asked GPT-3.5-Turbo\footnote{\scriptsize{\url{https://platform.openai.com/docs/models/gpt-3-5-turbo}}} to convert an existing code snippet into a more modular one, while ensuring its functional correctness.

However, we claim that their report warrants revisiting for two reasons.
First, since LLMs are notorious for their verbosity, it is unclear whether the conversion process aimed solely for modularity or accidentally introduced unexpected side effects.
Second, the lack of a formally defined quantitative method for estimating modularity hinders more extensive analyses related to the problem.

In this paper, we (re-)investigate the effectiveness of pursuing modularity in NL2Code generation.
We aim to push the boundaries of previous work by (1) introducing a novel metric that quantifies the modularity of a code snippet using numeric values. 
Based on the metric, we (2) classify code snippets as modular or non-modular without relying on LLMs, and evaluate how each category contributes to performance.\footnote{Note that this was infeasible in the previous study \cite{jain2024llmassisted} as there was no clear standard for determining whether each code snippet is modular or not.}
Moreover, beyond previous work, we (3) conduct experiments on models with parameters exceeding 7B (i.e., 33B and 34B) to investigate the impact of model scale.
Figure \ref{fig:main} illustrates the core research question of this work. 

In experiments, we discover that contrary to conventional wisdom in the literature, \textbf{the modularity of a code example may not be the crucial factor for performance}.
We also explore potential explanations for why LLMs do not exhibit a preference for modular code compared to non-modular one.

\begin{figure}[t!]
\centering
    \includegraphics[width=\columnwidth]{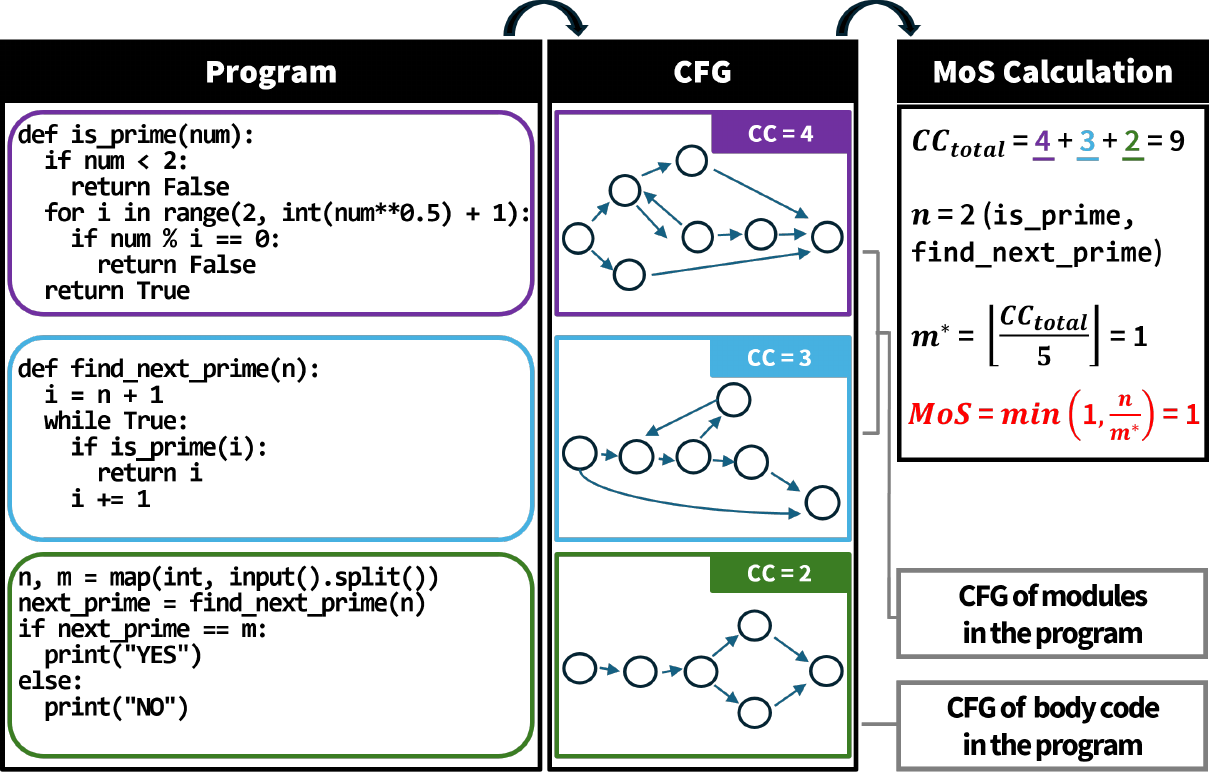}
    \caption{Procedure of computing Cyclomatic Complexity (CC) and Modularity Score ({\scshape MoS}). 
    We first build control-flow graphs (CFGs) from the given code to derive CC.
    The CC values are then used to compute {\scshape MoS} as the form of $\text{CC}_{\text{total}}$ and $m^*$.}
    \label{fig:MoS}
\end{figure}

\section{Quantitative Definition of Modularity}

To assess the impact of code modularity, the first essential step is to develop a method that provides a measurable score for code modularity.
While the previous study \cite{jain2024llmassisted} bypassed this vital step,\footnote{The authors instead utilized LLMs to transform all code snippets into supposedly modular ones.} we present a reasonable metric for estimating code modularity, which is challenging due to the inherent subjectivity of the concept itself.

Inspired by the software engineering literature, we employ the concept of \textbf{Cyclomatic Complexity (CC)} \cite{1702388} to determine the ideal number of modules, $m^*$, for a given code snippet. 
CC counts the number of independent execution paths in the control-flow graph (CFG) of the target code, where the CFG is a graph representation of all potential paths that a program might follow during execution.
CC can also be calculated as $E$ - $N$ + 2, where $E$ and $N$ correspond to the number of edges and nodes in the CFG.\footnote{In practice, we rely on the Python library Radon (\url{https://radon.readthedocs.io/)} to derive CC.}
The CC values are computed at either the whole code level (total CC; $\text{CC}_{\text{total}}$) or the function level (meaning the average CC across all functions in the code; $\text{CC}_{\text{avg}}$).

A high CC value generally indicates a complex code structure. 
It functions as a guideline for code decomposition, suggesting that a function whose CC is exceeding a certain threshold value $\tau$, e.g., 5 \cite{1702388} or 10 \cite{McConnell2004}, might benefit from being broken down into smaller sub-functions.
Based on the concept, we assume that the average CC of an ideal modular code example, denoted by $\text{CC}_{\text{avg}^*}$, should be equal to the threshold $\tau$.\footnote{Given two choices for $\tau$, i.e., 5 or 10, we set $\tau$ to 5 to encourage a sparser distribution of modularity scores ({\scshape MoS}).}
In other words, ideally, every function within a modular code snippet is expected to have a CC value of $\tau$.
Following the intuition, we define $m^*$, the number of ideal modules, as follows:

\begin{equation*}
m^* = \biggl \lfloor \frac{\text{CC}_{\text{total}}}{\text{CC}_{\text{avg}^*}} \biggr \rfloor = \biggl \lfloor \frac{\text{CC}_{\text{total}}}{\tau} \biggr \rfloor,
\end{equation*}

Finally, we define the modularity score, dubbed {\scshape MoS}, as follows:
\begin{equation*}
\text
{\scshape MoS} = \begin{cases}
    \min\left(1, \frac{n}{m^*}\right) & \text{if } {m^*} > 0 \\
    0 & \text{if } {m^*}=n=0 \\
    1 & \text{if } {m^*} = 0 {,\,n} > 0
\end{cases},
\end{equation*}
where $n$ is equal to the actual number of modules in the target code.\footnote{We consider modules valid only if they are utilized in at least one execution path of the program.} That is, the closer $n$ (actual number of modules) is to $m^*$ (ideal number of modules), the higher the modularity is considered to be.\footnote{In extreme cases where $m^*=0$ (no modularization required), the modularity score is set to 0 if no actual modules are used ($n=0$) and 1 otherwise ($n>0$).} 
The process of deriving {\scshape MoS} is illustrated in Figure \ref{fig:MoS}. 

In Appendix \ref{app:The Effectiveness of MoS}, we show that {\scshape MoS} is effective not only in capturing the structural properties of a code snippet but also in revealing the frequency of interactions between functions within the code.
\begin{table*}[t!]
\begin{center}\small
\begin{tabular}{lcccccccccc}
\toprule[1.5pt]

\multirow{2}[2]{*}{\textbf{Model}} & \multirow{2}[2]{*}{\textbf{Size}} & \multirow{2}[2]{*}{\shortstack[c]{\textbf{Code} \\ \textbf{Type}}} & \multicolumn{2}{c}{\textbf{Introductory}} & \multicolumn{2}{c}{\textbf{Interview}} & \multicolumn{2}{c}{\textbf{Competition}} & \multicolumn{2}{c}{\textbf{Average}} \\
\cmidrule(lr){4-5}\cmidrule(lr){6-7}\cmidrule(lr){8-9}\cmidrule(lr){10-11}
& & & pass@1 & pass@5 & pass@1  & pass@5 & pass@1 & pass@5  & pass@1 & pass@5 \\
\midrule
\multirow{4}{*}{Code Llama} & \multirow{4}{*}{7B}
& \textbf{MC} & 7.98 & 12.75 & 1.26 & 2.63 & 0.00 & 0.03 & 2.43 & 4.33 \\
& & \textbf{SC} & 11.12 & 15.78 & 1.65 & 3.13 & 0.07 & 0.26 & 3.32 & 5.29 \\
& & \cellcolor{gray!25}\makebox[0pt]{\textbf{TMC}} & \cellcolor{gray!25}\makebox[0pt]{\textbf{14.67}} & \cellcolor{gray!25}\makebox[0pt]{\textbf{19.63}} & \cellcolor{gray!25}\makebox[0pt]{\textbf{2.28}} & \cellcolor{gray!25}\makebox[0pt]{\textbf{3.98}} & \cellcolor{gray!25}\makebox[0pt]{\textbf{0.21}} & \cellcolor{gray!25}\makebox[0pt]{\textbf{0.59}} & \cellcolor{gray!25}\makebox[0pt]{\textbf{4.45}} & \cellcolor{gray!25}\makebox[0pt]{\textbf{6.66}} \\
& & \cellcolor{gray!25}\makebox[0pt]{\textbf{TSC}} & \cellcolor{gray!25}\makebox[0pt]{13.84} & \cellcolor{gray!25}\makebox[0pt]{17.15} & \cellcolor{gray!25}\makebox[0pt]{2.16} & \cellcolor{gray!25}\makebox[0pt]{3.61} & \cellcolor{gray!25}\makebox[0pt]{0.07} & \cellcolor{gray!25}\makebox[0pt]{0.24} & \cellcolor{gray!25}\makebox[0pt]{4.20} & \cellcolor{gray!25}\makebox[0pt]{6.07} \\
\midrule
\multirow{4}{*}{DeepSeekCoder} & \multirow{4}{*}{6.7B} 
    &\textbf{MC} & 24.76 & 32.59 & 6.49 & 10.58 & 0.72 & 1.72 & 9.39 & 14.01 \\
& & \textbf{SC} & 28.93 & 36.26 & 7.17 & 11.02 & 0.65 & 1.42 & 10.74 & 14.99 \\
& & \cellcolor{gray!25}\makebox[0pt]{\textbf{TMC}} & \cellcolor{gray!25}\makebox[0pt]{\textbf{34.26}} & \cellcolor{gray!25}\makebox[0pt]{\textbf{40.74}} & \cellcolor{gray!25}\makebox[0pt]{\textbf{9.60}} & \cellcolor{gray!25}\makebox[0pt]{\textbf{13.41}} & \cellcolor{gray!25}\makebox[0pt]{\textbf{0.76}} & \cellcolor{gray!25}\makebox[0pt]{\textbf{1.93}} & \cellcolor{gray!25}\makebox[0pt]{\textbf{13.49}} & \cellcolor{gray!25}\makebox[0pt]{\textbf{17.63}} \\
& & \cellcolor{gray!25}\makebox[0pt]{\textbf{TSC}} & \cellcolor{gray!25}\makebox[0pt]{33.24} & \cellcolor{gray!25}\makebox[0pt]{39.73} & \cellcolor{gray!25}\makebox[0pt]{8.55} & \cellcolor{gray!25}\makebox[0pt]{12.40} & \cellcolor{gray!25}\makebox[0pt]{0.55} & \cellcolor{gray!25}\makebox[0pt]{1.21} & \cellcolor{gray!25}\makebox[0pt]{12.55} & \cellcolor{gray!25}\makebox[0pt]{16.64} \\

\bottomrule[1.5pt]
\end{tabular}
\end{center}
\caption{Results on APPS measured by pass@$k$. 
We use $n$ = 10 for pass@1 and pass@5. 
The best results are in \textbf{bold} for each section. 
Two-shot prompting is applied for generating code given natural language queries.
}

\label{tab:experiment1-1}
\end{table*}

\begin{table}[t!]
\begin{center}\small
\begin{tabular}{llccc}
\toprule[1.5pt]
\multirow{2}[2]{*}{\textbf{Model}} & \multirow{2}[2]{*}{\textbf{Size}} & \multirow{2}[2]{*}{\shortstack[c]{\textbf{Code} \\ \textbf{Type}}} & \multicolumn{2}{c}{\textbf{CodeContests}} \\
\cmidrule(lr){4-5}
&  &  & pass@1 & pass@10 \\
\midrule
\multirow{8}{*}[-0.5ex]{Code Llama} & \multirow{4}{*}{7B} 
& \textbf{MC} & 1.98 & 8.02 \\
&  & \textbf{SC} & 2.58 & 8.81 \\
&  & \cellcolor{gray!25}\makebox[0pt]{\textbf{TMC}} & \cellcolor{gray!25}\makebox[0pt]{2.57} & \cellcolor{gray!25}\makebox[0pt]{10.18}\\
&  & \cellcolor{gray!25}\makebox[0pt]{\textbf{TSC}} & \cellcolor{gray!25}\makebox[0pt]{\textbf{4.35}} & \cellcolor{gray!25}\makebox[0pt]{\textbf{10.67}} \\
\cmidrule{2-5}
& \multirow{4}{*}{34B} 
& \textbf{MC} & 4.11 & 12.78 \\
& & \textbf{SC} & \textbf{5.83} & 14.1 \\
& & \cellcolor{gray!25}\makebox[0pt]{\textbf{TMC}} & \cellcolor{gray!25}\makebox[0pt]{3.39} & \cellcolor{gray!25}\makebox[0pt]{13.55} \\
& & \cellcolor{gray!25}\makebox[0pt]{\textbf{TSC}} & \cellcolor{gray!25}\makebox[0pt]{5.61}& \cellcolor{gray!25}\makebox[0pt]{\textbf{15.32}} \\
\midrule
\multirow{8}{*}[-0.5ex]{DeepSeekCoder} & \multirow{4}{*}{6.7B} 
& \textbf{MC} & 5.3 & 12.78  \\
& & \textbf{SC} & 7.15  & 16.27 \\ 
& & \cellcolor{gray!25}\makebox[0pt]{\textbf{TMC}} & \cellcolor{gray!25}\makebox[0pt]{8.02}& \cellcolor{gray!25}\makebox[0pt]{\textbf{17.88}} \\
& & \cellcolor{gray!25}\makebox[0pt]{\textbf{TSC}} & \cellcolor{gray!25}\makebox[0pt]{\textbf{8.19}}& \cellcolor{gray!25}\makebox[0pt]{17.79}\\
\cmidrule{2-5}
& \multirow{4}{*}{33B} 
& \textbf{MC} & 6.79 & 16.14 \\
& & \textbf{SC} & 8.87 & 20.5 \\
& & \cellcolor{gray!25}\makebox[0pt]{\textbf{TMC}} & \cellcolor{gray!25}\makebox[0pt]{\textbf{9.38}} & \cellcolor{gray!25}\makebox[0pt]{\textbf{22.74}}\\
& & \cellcolor{gray!25}\makebox[0pt]{\textbf{TSC}} & \cellcolor{gray!25}\makebox[0pt]{8.78}& \cellcolor{gray!25}\makebox[0pt]{22.09}\\
\midrule
\multirow{4}{*}[-0.5ex]{GPT-4o-mini} & \multirow{4}{*}{-}
& \textbf{MC} & 14.07 & -\\
& & \textbf{SC} & \textbf{15.35} & - \\
& & \cellcolor{gray!25}\makebox[0pt]{\textbf{TMC}} & \cellcolor{gray!25}\makebox[0pt]{14.29}& - \\
& & \cellcolor{gray!25}\makebox[0pt]{\textbf{TSC}} & \cellcolor{gray!25}\makebox[0pt]{14.4}& - \\
\bottomrule[1.5pt]
\end{tabular}
\end{center}
\caption{Results on CodeContests measured by pass@$k$. We use $n$ = 10 for pass@1 and $n$ = 50 for pass@10, respectively. The best results are in \textbf{bold} for each section. 
Two-shot prompting is applied for generating code given natural language queries. 
Due to the cost issue, we only compute pass@1 for GPT-4o-mini.
}
\label{tab:experiment1-2}
\end{table}

\section{Four Code Categories by Modularity}\label{sec:Four Code Categories by Modularity}

With a way to quantify code modularity, we can now classify a code dataset into two categories---modular and non-modular\,(=\,singular).
We further leverage prior research by including LLM-based code transformations and their corresponding manually recovered counterparts for controlled experiments. 
This allows us to create four distinct clusters of code separated by their modularity levels.\footnote{Figure \ref{fig:4code_example} in Appendix displays examples of each category.}

\hfill \break
\textbf{Modular Code (MC)} is a collection of code snippets with high {\scshape MoS} among solutions for each problem in a dataset.

\hfill \break
\textbf{Singular Code (SC)} represents another set of solution code examples for the same problems corresponding to \textbf{MC}, with {\scshape MoS} being 0.

\hfill \break
\textbf{Transformed Modular Code (TMC)} can be obtained by utilizing GPT-3.5-Turbo ($f$) to transform \textbf{SC} into code with high {\scshape MoS} while preserving its original functionality.
The conversion process can be represented by the following: $$\textbf{TMC}= f({I},{Q},\textbf{SC}),$$ where $I$ represents a transformation instruction and $Q$ is the problem description of \textbf{SC}.\footnote{See Figure \ref{fig:prompt_transformation} for prompt details on the conversion process.}

\hfill \break
\textbf{Transformed Singular Code (TSC)} is a variation from \textbf{TMC}, whose modularity is manually removed by human programmers.
The goal of this approach is to ensure that all factors except modularity are preserved during the conversion process from \textbf{TMC} to \textbf{TSC}, minimizing unintended changes that could occur if we rely solely on automatic conversion.

Specifically, \textbf{TSC} is created by replacing the module invocation parts in \textbf{TMC} with the body of the corresponding modules and then removing those modules from the program.
By comparing \textbf{TSC} and \textbf{TMC}, which are expected to be identical except for their modularity, we gain a valuable opportunity to rigorously assess the impact of modularity while accounting for the influence of the transformation process executed by $f$.

\section{Experimental Setups}

We explore the impact of modularity by comparing how the four code collections, categorized by their modularity levels, affect performance.
We first focus on the case of utilizing code LLMs with few-shot in-context learning.
We leverage two-shot demonstrations (providing two code examples) unless otherwise specified.\footnote{Refer to Figure \ref{fig:prompt_codellama_figure}, \ref{fig:prompt_deepseek_figure}, and
\ref{fig:prompt_gpt_figure} for prompt details.}
In addition, we explore the scenario of fine-tuning LLMs with datasets that have varying levels of modularity.

\paragraph{Models} We exploit three LLMs for code generation---Code Llama (7B, 34B; \citealp{roziere2024code}), DeepSeekCoder (6.7B, 33B; \citealp{guo2024deepseekcoder}), and GPT-4o-mini.\footnote{\scriptsize{\url{https://openai.com/index/gpt-4o-mini-advancing-cost-efficient-intelligence/}}}

\paragraph{Datasets}

We employ two NL2Code generation datasets---APPS \cite{hendrycks2021measuring} and CodeContests \cite{Li_2022}.\footnote{Note that representative code generation benchmarks, e.g., HumanEval \cite{chen2021evaluating}, typically provide code snippets whose length restricts the possibility of modularization.}
They are based on competitive programming contests and provide a set of different solutions for each problem.\footnote{We preprocess the APPS and CodeContests datasets following \citet{jain2024llmassisted}. Refer to Appendix \ref{app:Dataset Preprocessing} for more details.}
In this study, we focus our evaluation on Python.

For ICL experiments, \textbf{MC} and \textbf{SC} demonstrations are chosen from solutions for random problems sampled from each dataset. 
The one with the highest {\scshape MoS} among solutions is chosen as \textbf{MC}.
After selecting \textbf{SC} examples, they are converted into \textbf{TMC}, and finally, \textbf{TSC} is manually derived.

For fine-tuning, we split the original dataset into two subsets, \textbf{MC} and \textbf{SC}, and train different variations of LLMs on each subset. 
The details on fine-tuning experiments are presented in Appendix \ref{app:Training Details}.

\paragraph{Evaluation Metrics}

We apply an unbiased version of pass@$k$ \cite{chen2021evaluating}, which measures the functional correctness of generated programs by running them against test cases. 
For each problem, LLMs are prompted to generate $n$ programs, and we determine $c$, the number of programs that pass the test cases.
In addition, $k$ ($k \leq n$) specifies the granularity of evaluation such that the metric indicates the probability of finding at least one correct solution when sampling $k$ programs out of the $n$ generated ones.
The metric is then averaged over all problems.
As a result, pass@$k$ is computed as:
\begin{equation*}
\text{pass}@k = \mathbb{E}_{\text{problems}}\left[1-\frac{\binom{n-c}{k}}{\binom{n}{k}}\right].
\end{equation*}

\section{Main Results}
\label{Main Results}

Table \ref{tab:experiment1-1} and Table \ref{tab:experiment1-2} present the results of experiments conducted in the ICL setting on APPS and CodeContests, categorized by the modularity of the code demonstrations.
All results are the average of five independent runs with different random seeds.

In Table \ref{tab:experiment1-1}, we observe that \textbf{SC} outperforms \textbf{MC}, but as previously reported, the performance of \textbf{TMC} is slightly better than \textbf{TSC}.
However, their marginal performance gaps raise questions about the impact of modularity.

In Table \ref{tab:experiment1-2}, the relationship between modularity and performance becomes less clear.
When comparing \textbf{MC} to \textbf{SC}, we observe that \textbf{MC} consistently underperforms \textbf{SC}, which contradicts previous findings.
Furthermore, the comparison between \textbf{TMC} and \textbf{TSC}---a more controlled setting for evaluating modularity---shows no clear correlation between code modularity and performance. 
This is despite the fact that the transformation process by GPT-3.5-Turbo (\textbf{SC} $\rightarrow$ \textbf{TMC}) seems to contribute to non-trivial increases in performance, particularly for Code Llama and DeepSeekCoder.
GPT-4o-mini demonstrates consistent performance across all four code types, suggesting that modularity does not significantly impact its performance.

We thus argue that the previously reported effectiveness of modularity on performance was likely due to unforeseen consequences of the transformation process, rather than the modularity itself. 

On the other hand, the performance of LLMs fine-tuned on \textbf{MC} and \textbf{SC} from CodeContests is reported in Table \ref{tab:ft}. 
We discover that \textbf{SC} constantly outperforms \textbf{MC}, albeit by a narrow margin, reflecting a similar trend observed in the ICL setting.\footnote{Due to the cost of constructing \textbf{TMC} and \textbf{TSC} using GPT-3.5-Turbo, we focused our experiments on \textbf{MC} and \textbf{SC}.} 
These results suggest that the modularity of the code examples used for training does not have a significant impact on the performance of LLMs in terms of code generation.

\begin{table}[t]
\begin{center}\small
\begin{tabular}{lcccc}
\toprule[1.5pt]
\multirow{2}[2]{*}[-0.2ex]{\textbf{Model}} & \multirow{2}[2]{*}[-0.2ex]{\textbf{Size}} & \multirow{2}[2]{*}{\shortstack[c]{\textbf{Code} \\ \textbf{Type}}} & \multicolumn{2}{c}{\textbf{CodeContests}} \\
\cmidrule(lr){4-5}
& & & pass@1 & pass@10 \\
\midrule
\multirow{2}{*}[-0.2ex]{Code Llama} & \multirow{2}{*}[-0.2ex]{7B}  & \textbf{MC} & 3.88 & 12.2 \\
& & \textbf{SC} & \textbf{4.42} & \textbf{12.56} \\
\midrule
\multirow{2}{*}[-0.2ex]{DeepSeekCoder} & \multirow{2}{*}[-0.2ex]{6.7B} & \textbf{MC} & 6.06 & 13.82 \\
& & \textbf{SC} & \textbf{8.73} & \textbf{16.16} \\
\bottomrule[1.5pt]
\end{tabular}
\end{center}
\caption{Performance of fine-tuning code LLMs on CodeContests, measured by pass@$k$. We use $n$ = 10 for pass@1 and $n$ = 50 for pass@10. The best results in each section are highlighted in \textbf{bold}. Zero-shot prompting is used during inference, meaning no demonstrations are provided to guide the models in generating code.}
\label{tab:ft}
\end{table}

\section{Analysis}

\begin{table}[t]
\begin{center}\small
\begin{tabular}{lccc}
\toprule[1.5pt]
\textbf{Model} & \textbf{Size} & \textbf{Pearson} & \textbf{Spearman} \\
\midrule
Code Llama & 7B & -0.34 (0) & -0.31 (0) \\
\midrule
DeepSeekCoder & 6.7B & -0.21 (0.04) & -0.25 (0.01) \\
\bottomrule[1.5pt]
\end{tabular}
\caption{Correlations between modularity ({\scshape MoS}) and performance (pass@1), evaluated on CodeContests. 
They consistently show weak negative relationships.
Numbers in parentheses represent $p$-values.}
\label{tab:corr}
\end{center}
\end{table}

\subsection{Correlation Study}

We conduct an extra experiment to dive deeper into the modularity-performance relationship.
Specifically, given 100 code samples used as demonstrations,\footnote{For balanced sampling, we create bins along the {\scshape MoS} dimension and sample an equal number of data from each bin. All the examples are either \textbf{MC} or \textbf{SC} type.} we compute the Pearson and Spearman correlations between their modularity ({\scshape MoS}) and resulting performance (pass@1).
For simplicity, we perform one-shot ICL on CodeContests.
Experimental results are presented in Table \ref{tab:corr} and Figure \ref{fig:corr} in Appendix.
Surprisingly, the results reveal weak negative correlations between modularity and performance, suggesting that modularity may not offer benefits, or even hinder performance in some cases.

\subsection{Do LLMs Favor Modular Code?}
The minimal performance gap between \textbf{(T)MC} and \textbf{(T)SC} suggests that LLMs may not have a strong preference for generating modular code.
To verify this hypothesis, we compare the perplexities of LLMs on modular and non-modular code.
Formally, the perplexity of a code snippet $\mathcal{C}$ given a problem description $\mathcal{D}$ is:
$$PPL(\mathcal{C})=exp\biggl\{-\frac{1}{n}\sum_{t=0}^{n-1} log{P(x_{t+1}\,|\,\mathcal{D},x_{\leq t})}\biggr\},$$
where $\mathcal{C}$, consisting of tokens $x_{1},\dots,x_{n}$, belongs to either \textbf{MC} ($\mathcal{C}_{\text{MC}}$) or \textbf{SC} ($\mathcal{C}_{\text{SC}}$).
We sample nearly 10,000 problems from CodeContests containing both $\mathcal{C}_{\text{MC}}$ and $\mathcal{C}_{\text{SC}}$, with $\mathcal{C}_{\text{MC}}$ having {{\scshape MoS}} values ranging from 0.7 to 1 and $\mathcal{C}_{\text{SC}}$ having value of 0.
We then compare $PPL(\mathcal{C}_{\text{MC}})$ and $PPL(\mathcal{C}_{\text{SC}})$ averaged over all examples to identify which type of code is better predicted by code language models.

Table \ref{tab:probexp} supports our hypothesis, highlighting a neutral preference of LLMs which is not biased towards generating \textbf{SC} or \textbf{MC}. 
This is presumably because the models were likely exposed to both code types during pre-training.
We speculate that this could be one of the reasons why modular examples are not always beneficial for code generation in language models.

\begin{table}[t!]
\begin{center}\small
\begin{tabular}{lccc}
\toprule[1.5pt]
\textbf{Model} & \textbf{Size} & $PPL(\mathcal{C}_\textbf{MC})$ & $PPL(\mathcal{C}_\textbf{SC})$ \\
\midrule
\multirow{2}[2]{*}{Code Llama} & 7B & 2.2 (0.57) & 2.4 (1) \\
\cmidrule{2-4}
& 34B & 2.02 (0.45) & 2 (0.44) \\
\midrule
\multirow{2}[2]{*}{DeepSeekCoder}& 6.7B & 1.93 (0.41) & 2.05 (0.63)\\
\cmidrule{2-4}
& 33B & 1.89 (0.42) & 1.89 (0.42) \\
\bottomrule[1.5pt]
\end{tabular}
\caption{Perplexities of LLMs for $\mathcal{C}_{\text{MC}}$ and $\mathcal{C}_{\text{SC}}$. LLMs exhibit similar predictive ability for both \textbf{SC} and \textbf{MC}. Numbers in parentheses represent standard deviations.}
\label{tab:probexp}
\end{center}
\end{table}

\section{Conclusion}

In this work, we propose a metric, called {\scshape MoS}, for quantifying the modularity of code snippets and evaluate its impact on performance. 
Our evaluation reveals no significant correlation, or even a possible weak negative correlation, between modularity and performance. 
This suggests that factors influencing the usefulness of code examples may differ between human and LLM perspectives. Exploring the influence of other code properties beyond modularity is a promising direction for future work.
 
\section*{Limitations}

Due to limited computational resources, we focused on designing experimental settings that are both targeted and generalizable. 
This limitation restricted the scope of our investigation, but considering more extensive configurations in future work---such as employing much larger models, and evaluating other programming languages---will help validate and potentially broaden the applicability of our findings.
Despite these limitations, we believe our findings offer valuable insights, thanks to our comprehensive exploration of the feasible configurations within the available resources. Additionally, identifying a core factor besides modularity that directly affects performance holds significant promise for improving code generation.

\section*{Ethics Statement}
In this study, we utilize models and datasets publicly available on Huggingface, ensuring that no
ethical issues are associated with their usage. 
All datasets for evaluation are open-source and follow strictly to data usage policies. 

\section*{Acknowledgements}
This work was supported by Institute of Information \& communications Technology Planning \& Evaluation (IITP) grant funded by the Korea government (MSIT) (No.RS-2020-II201373, Artificial Intelligence Graduate School Program (Hanyang University)), Institute of Information \& communications Technology Planning \& Evaluation (IITP) under the artificial intelligence semiconductor support program to nurture the best talents (IITP-2024-RS-2023-00253914) grant funded by the Korea government (MSIT), and the National Research Foundation of Korea (NRF) grant funded by the Korea government (MSIT) (No.2018R1A5A7059549).

\bibliography{custom}


\appendix
\section*{Appendix}
\label{sec:appendix}

\section{The Effectiveness of {\scshape MoS}}
\label{app:The Effectiveness of MoS}
In this section, we explore the question: Does {\scshape MoS} indeed reflect the modularity of code? A key aspect of modular programming is decomposing a program into modules and enabling interaction between them through function calls or class instantiation. We argue that {\scshape MoS} not only considers the structure of a program but also inherently reflects the interactions between its constituents (e.g., functions and class methods).

To support this claim, we examine the correlation between {\scshape MoS} and the number of function calls to determine if {\scshape MoS} reflects module interactions. Specifically, we (1) sample 100 codes from the CodeContests training set, (2) calculate the Pearson and Spearman correlation coefficients between each code example's {\scshape MoS} and the frequency of the function invocations, and (3) repeat the same experiment using five random seeds. To ensure balanced sampling, we create bins based on {\scshape MoS} values and sample the same number of codes from each bin. As shown in Table \ref{tab:corrmosfc}, the Pearson and Spearman correlation coefficients are 0.41 and 0.61, respectively. This positive correlation between {\scshape MoS} and the frequency of function calls highlights its effectiveness in reflecting modular interaction.

\section{Dataset Preprocessing}
\label{app:Dataset Preprocessing}
In both the APPS and CodeContests datasets, there are some solution codes that are incorrect based on functional correctness. 
We filter out code snippets that do not pass the test cases and retain only the solutions written in Python. 
After data filtering, CodeContest has a training dataset of around 7K samples, while APPS has a training dataset of approximately 2K samples. 
Since some of the problems in APPS provide insufficient or absent test cases, we retain only problems obtained from \textsf{atcoder}, \textsf{codechef}, and \textsf{codeforces} in APPS, following \citet{jain2024llmassisted}. 
APPS are divided into APPS-INTRODUCTORY, APPS-INTERVIEW, and APPS-COMPETITION based on problem difficulty. 
Table \ref{tab:datastats} describes the statistics of the APPS and CodeContests datasets we finally employed. 
Additionally, we also guarantee that both \textbf{TMC} and \textbf{TSC} pass the test cases.

\section{Training Details}
\label{app:Training Details}
We construct two training datasets consisting solely of \textbf{MC} and \textbf{SC}, respectively, based on their {\scshape MoS} values, and fine-tune the full parameters of the code LLMs on these datasets.
Both training subsets cover the same set of problems, with \textbf{SC} having a {\scshape MoS} value of 0 and \textbf{MC} having {\scshape MoS} values ranging between 0.7 and 1.
Following \citet{jain2024llmassisted}, we employ minhash-based deduplication using Gaoya\footnote{\url{https://github.com/serega/gaoya}} to limit each problem to a maximum of 25 solutions. Then, an equal number of codes are randomly selected from each problem to ensure that the \textbf{MC} and \textbf{SC} datasets have the same number of training samples. Applying this process to the CodeContests dataset results in about 5K problems and 61K training examples for both the \textbf{MC} and \textbf{SC} datasets.
We decided to exclude the APPS dataset from the fine-tuning experiment due to the limited number of common problems and codes for \textbf{SC} and \textbf{MC}.

For model training, we used the HuggingFace Trainer\footnote{\url{https://huggingface.co/docs/transformers/main\_classes/trainer}} library with the AdamW\cite{loshchilov2019decoupledweightdecayregularization} optimizer, starting with a learning rate of 5e-5. A cosine learning rate scheduler with a warmup ratio of 0.01 was applied, and we utilized bf16 precision to optimize memory usage. The effective batch size was set to 64, achieved through a per-device batch size of 4 and gradient accumulation step of 16. Training was conducted for 1 epoch on 4 A6000 GPUs. After training, model inference was conducted in a zero-shot manner using the same sampling parameters as those in the ICL setting.

\begin{table}[t!]
\begin{center}\small
\begin{tabular}{ccc}
\toprule[1.5pt]
\textbf{Random Seed} & \textbf{Pearson} & \textbf{Spearman} \\
\midrule
27 & 0.37 (0) & 0.61 (0) \\
\midrule
42 & 0.41 (0) & 0.64 (0) \\
\midrule
101 & 0.4 (0) & 0.63 (0) \\
\midrule
134 & 0.44 (0) & 0.57 (0) \\
\midrule
169 & 0.42 (0) & 0.62 (0) \\
\midrule
\textbf{Average} & \textbf{0.41} & \textbf{0.61} \\
\bottomrule[1.5pt]
\end{tabular}
\caption{Correlations between {\scshape MoS} of code and number of function calls in the code. 
Numbers in parentheses represent p-values.}
\label{tab:corrmosfc}
\end{center}
\end{table}

\section{Details on In-Context Learning}
\label{app:Details on In-Context Learning}
\begin{figure*}[h]
\centering
    \includegraphics[width=\textwidth]{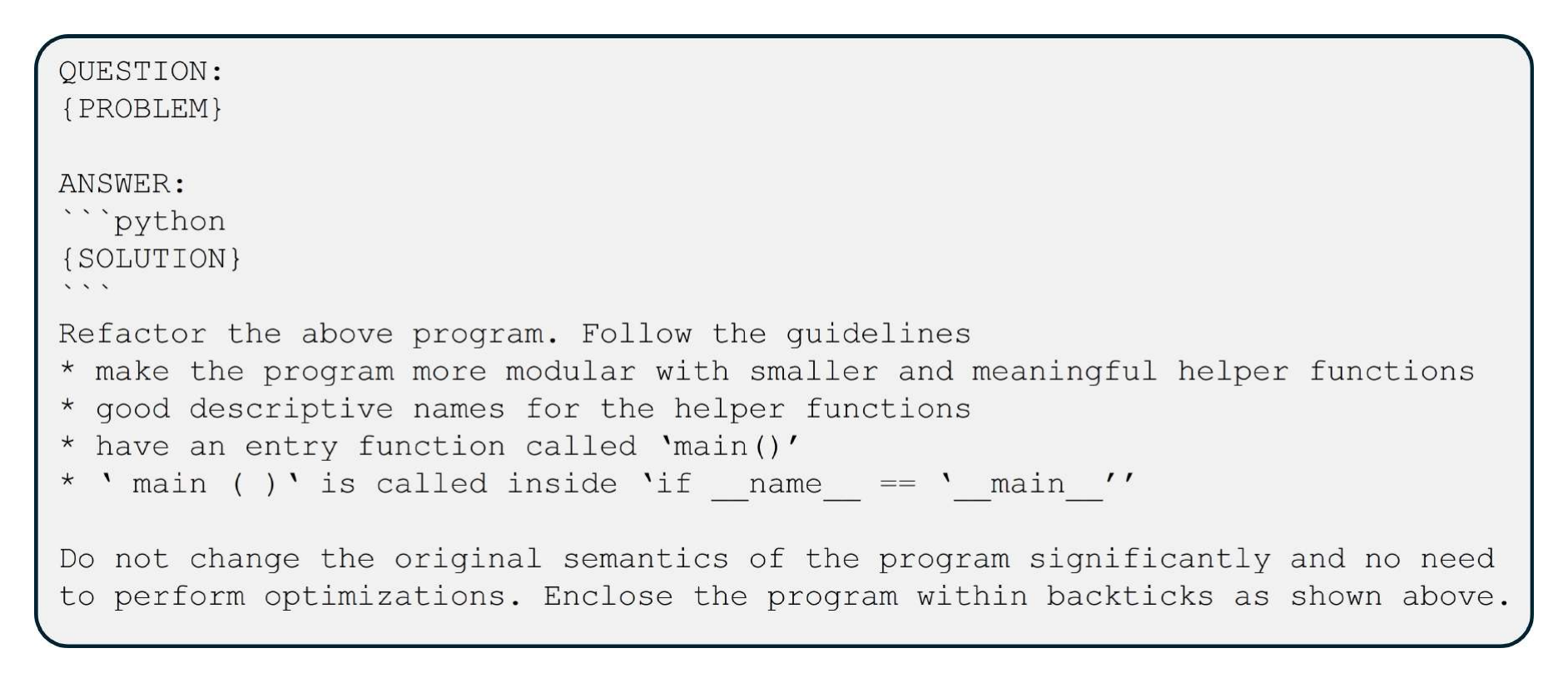}
    \caption{The prompt template used for converting \textbf{SC} to \textbf{TMC}.}
    \label{fig:prompt_transformation}
\end{figure*}

Following \citet{roziere2024code}, we use a special instruction to help models understand the specific question format: ``\textit{read from and write to standard IO}'' for standard questions and ``\textit{use the provided function signature}'' for call-based questions, which we insert into our prompt as the question guidance for APPS and use special instructions for standard questions for CodeContests.  This corresponds to \{FEW\_SHOT\_QUESTION\} in Figure \ref{fig:prompt_transformation}. 
We use temperature and top-p sampling strategies for calculating pass@$k$. 
Following \citet{jain2024llmassisted}, we set the top-p value of 0.95 and temperature to 0.1 for pass@$1$, and 0.6 for pass@$10$.



\clearpage
\begin{figure*}[t]
\centering
    \includegraphics[width=\textwidth]{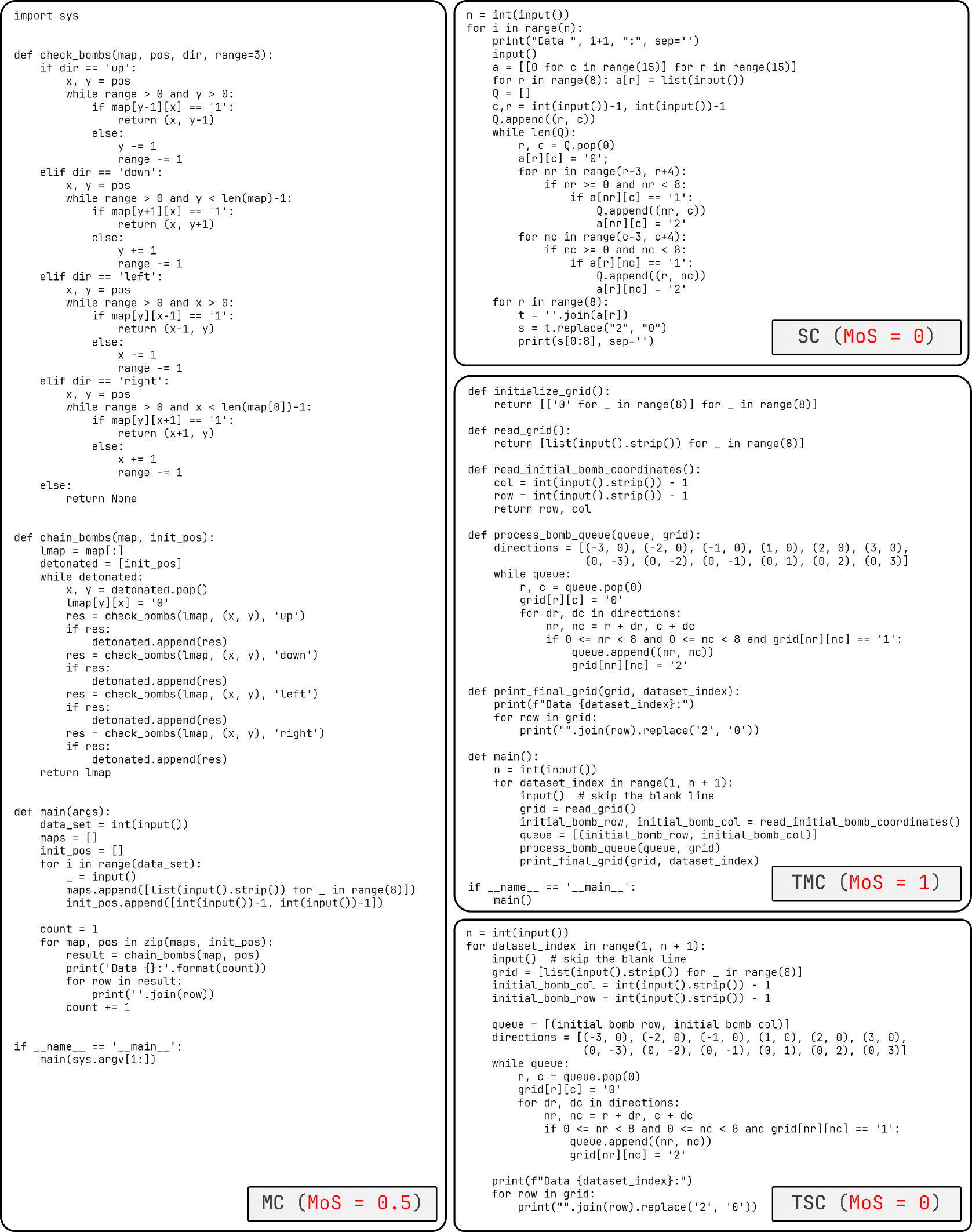}
    \caption{Examples of four code categories for the same problem with their modularity scores.}
    \label{fig:4code_example}
\end{figure*}

\begin{table*}[htbp]
\begin{center} 
\small
\begin{tabular}{lccccc} 
    \toprule[1.5pt] 
    & \multirow{2}{*}[-0.2ex]{\textbf{Split}}  & \multirow{2}{*}[-0.2ex]{\textbf{CodeContests}} & \multirow{2}{*}[-0.2ex]{\shortstack[c]{\textbf{APPS} \\ \textbf{(INTRODUCTORY)}}} & \multirow{2}{*}[-0.2ex]{\shortstack[c]{\textbf{APPS} \\ \textbf{(INTERVIEW)}}}  &  \multirow{2}{*}[-0.2ex]{\shortstack[c]{\textbf{APPS} \\ \textbf{(COMPETITION)}}} \\
    & & & & &  \\
    \midrule
    \multirow{2}{*}[-0.2ex]{\# Problems} & Training & 7313 & 42 & 1247 & 361  \\
    \cmidrule{2-6}
    & Test & 165 & 702 & 2699 & 309 \\
    \midrule
    \multirow{2}{*}[-0.2ex]{\# Avg. Test Cases}& Training & 20 & 1 & 1 & 10 \\
    \cmidrule{2-6}
    & Test & 10 & 16 & 24 & 45 \\
    \midrule
    \# Avg. Solutions & Training & 182 & 64 & 24 & 17 \\
    \bottomrule[1.5pt]
\end{tabular}
\caption{Statistical details regarding the number of problems, the average number of test cases per problem, and the average number of solutions in the filtered datasets of CodeContests and APPS.}
\label{tab:datastats}
\end{center}
\end{table*}

\begin{figure*}[h]
\centering
    \includegraphics[width=\textwidth]{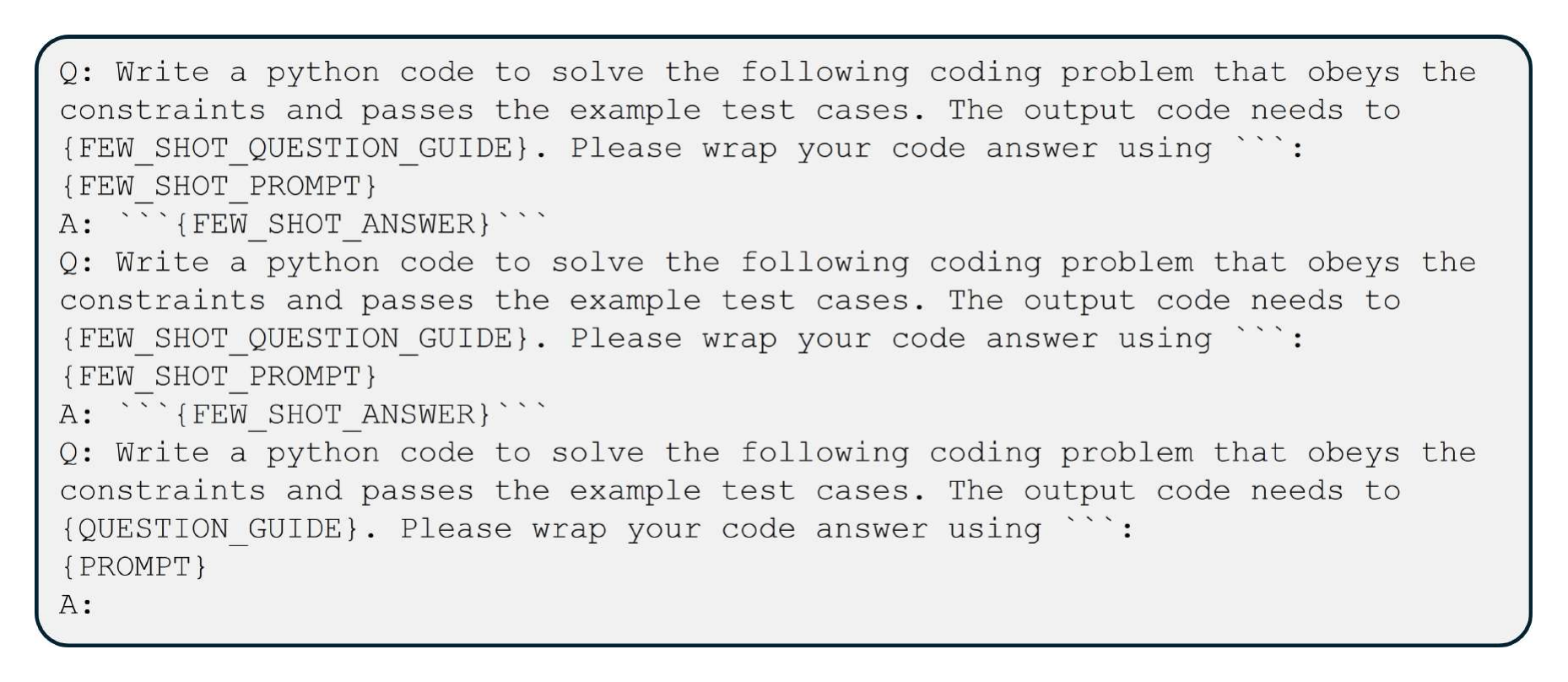}
    \caption{The prompt template used for two-shot in-context learning with Code Llama.}
    \label{fig:prompt_codellama_figure}
\end{figure*}

\begin{figure*}[htbp]
\centering
    \includegraphics[width=\textwidth]{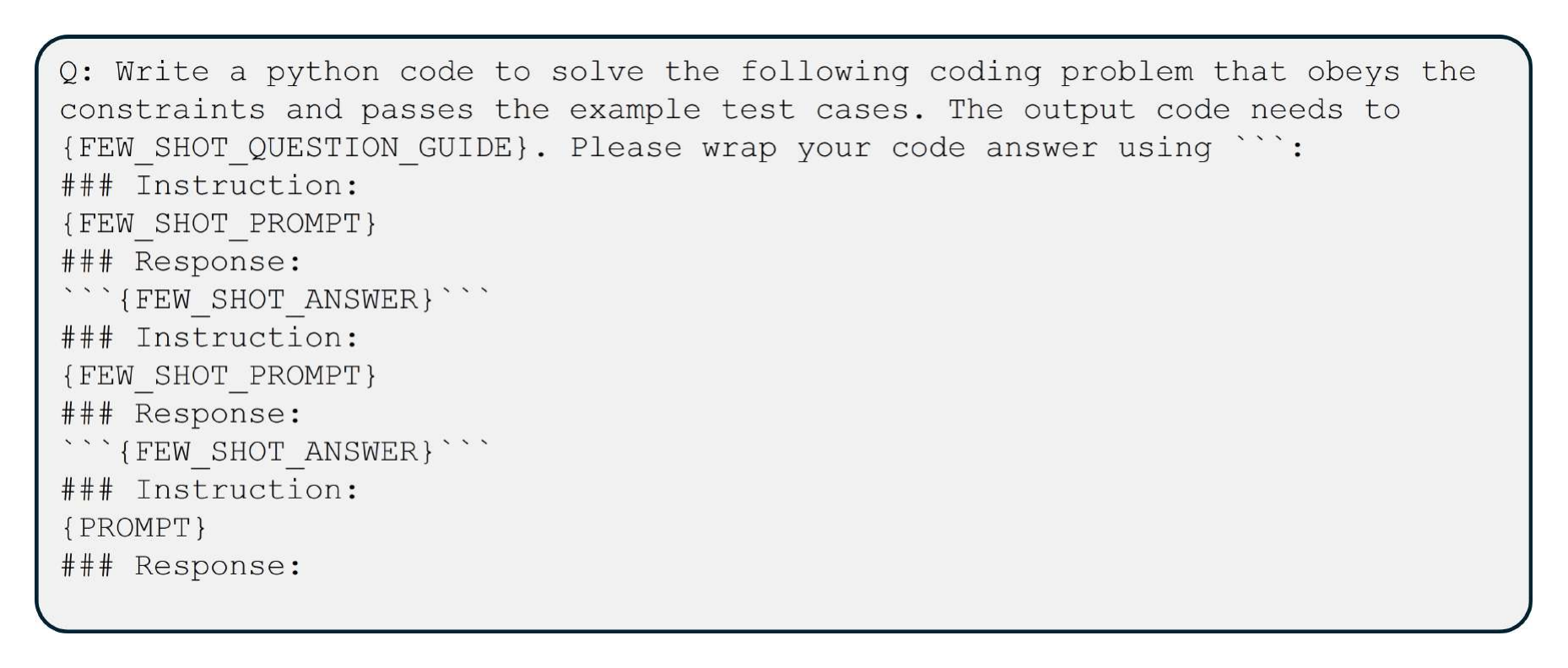}
    \caption{The prompt template used for two-shot in-context learning with DeepSeekCoder.}
    \label{fig:prompt_deepseek_figure}
\end{figure*}

\begin{figure*}[htbp]
\centering
    \includegraphics[width=\textwidth]{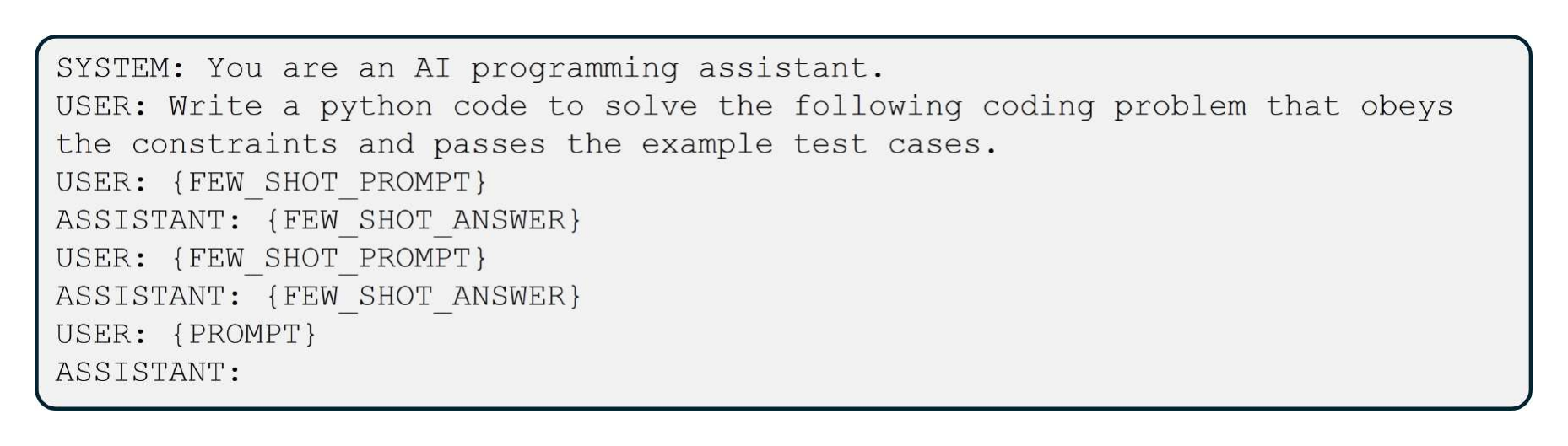}
    \caption{The prompt template used for two-shot in-context learning with GPT-4o-mini.}
    \label{fig:prompt_gpt_figure}
\end{figure*}

\begin{figure*}[htbp]
    \centering
    \begin{subfigure}[b]{0.49\textwidth}
        \centering
        \includegraphics[width=\textwidth]{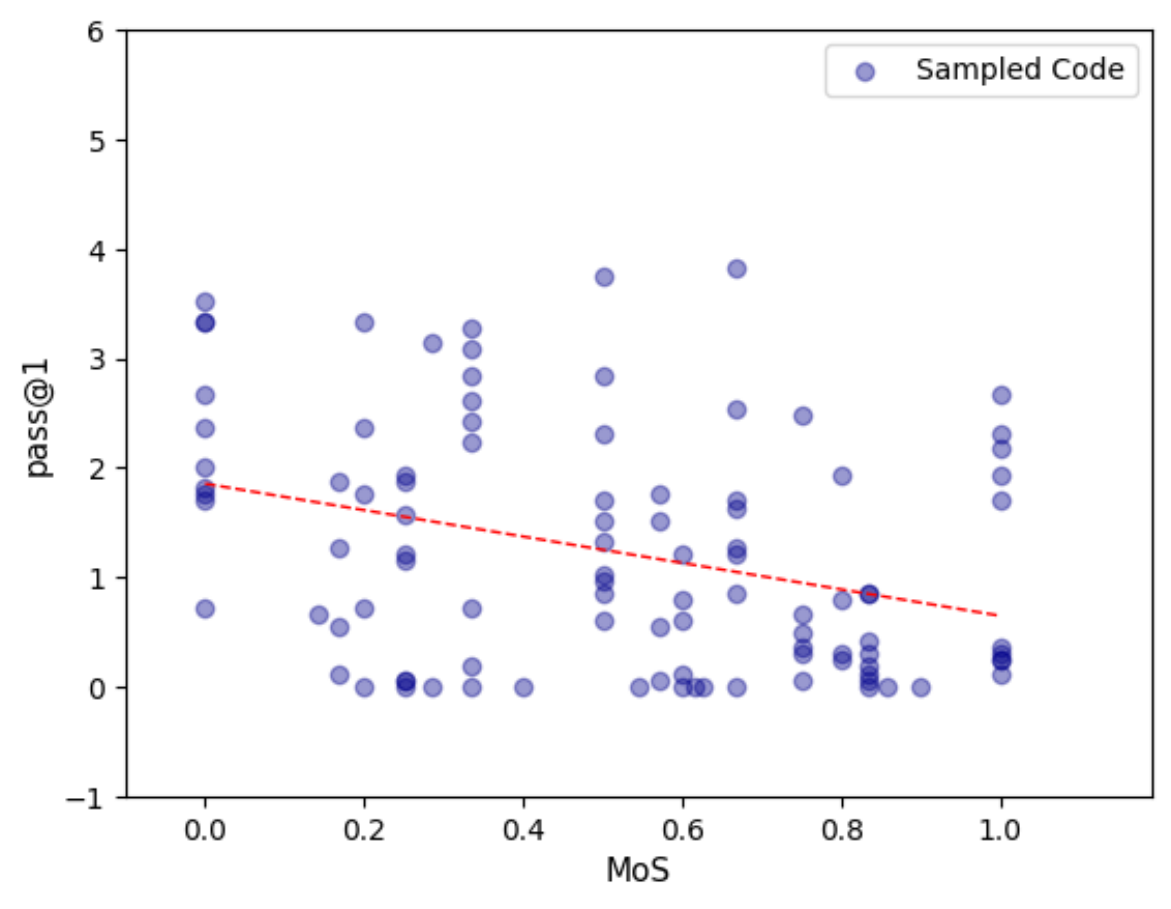}
        \caption{One-shot ICL with CodeLlama 7B.}
        \label{fig:corr(codellama)}
    \end{subfigure}
    \hspace{0.005\textwidth}
    \begin{subfigure}[b]{0.49\textwidth}
        \centering
        \includegraphics[width=\textwidth]{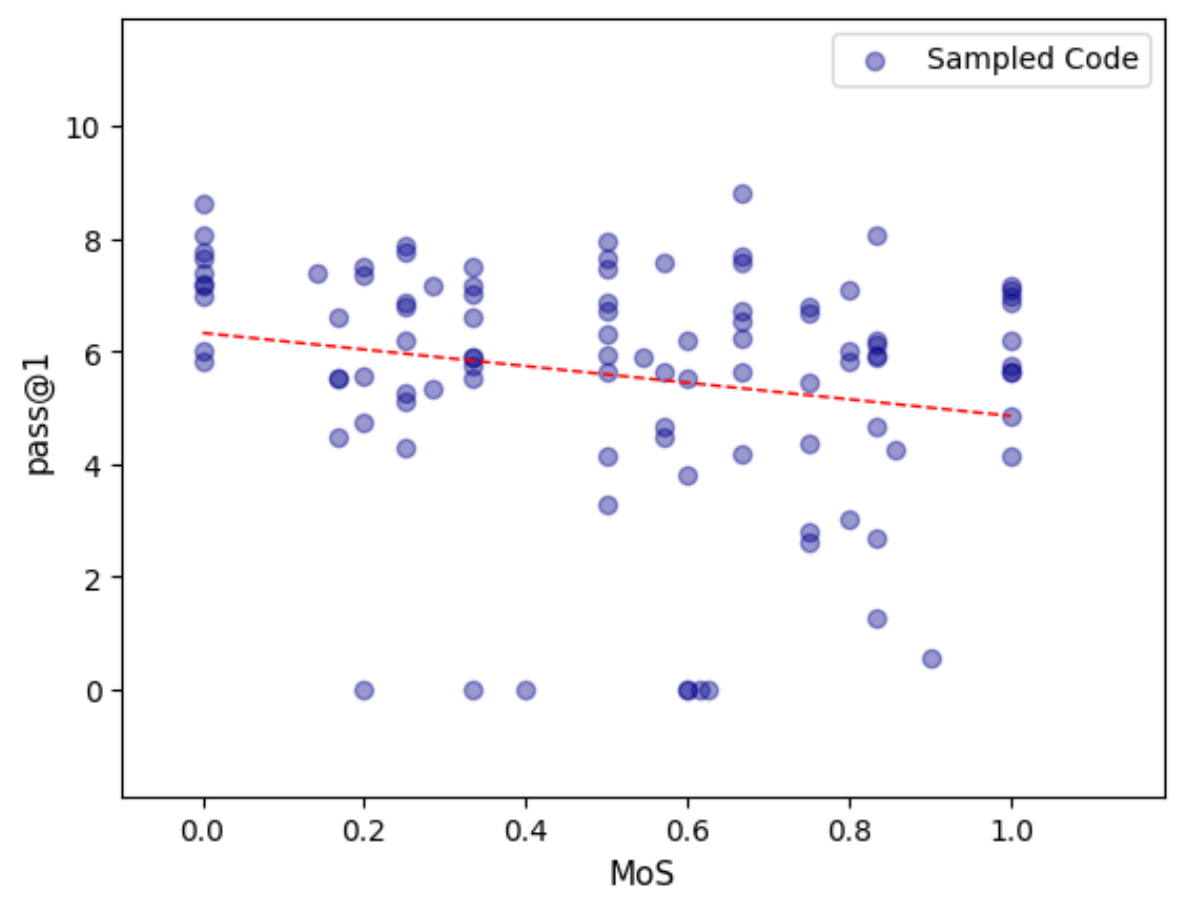}
        \caption{One-shot ICL with DeepSeekCoder 6.7B.}
        \label{fig:corr(deepseek)}
    \end{subfigure}
    \caption{Scatter plots with modularity ({\scshape MoS}) on the x-axis and performance (pass@1) on the y-axis show weak negative correlations between the two variables. 
    The CodeContests dataset is used for evaluation. 
    Note that the {\scshape MoS} scores of demonstration codes exhibit a wide distribution. 
    The red dashed line represents the regression line.}
    \label{fig:corr}
\end{figure*}

\end{document}